\pdfoutput=1

\documentclass[11pt]{article}

\usepackage[preprint]{acl}

\usepackage{times}
\usepackage{latexsym}
\usepackage{listings}

\usepackage[T1]{fontenc}

\usepackage[utf8]{inputenc}

\usepackage{microtype}

\usepackage{inconsolata}

\usepackage{graphicx}
\usepackage{amsmath}
\usepackage{amssymb}
\usepackage{graphicx}
\usepackage{tabularx}
\usepackage{multirow}
\usepackage{multicol}
\usepackage[justification=justified,singlelinecheck=false]{caption}
\usepackage{ragged2e}
\usepackage[ruled,vlined,linesnumbered]{algorithm2e}
\usepackage{verbatim}
\usepackage{booktabs}
\usepackage{fontawesome5}
\usepackage{subcaption}
\usepackage[draft]{minted}
\usepackage{hyperref}
\usepackage[noabbrev,capitalize,nameinlink]{cleveref}
\usepackage{tikz}
\usepackage{tikz-dependency}
\usepackage{adjustbox}
\usepackage{todonotes}
\usepackage[multiple]{footmisc}
\usepackage{pgfplots}
\usepackage{rotating}
\usetikzlibrary{positioning, fadings}
\usepackage{transparent}
\usepackage{cclicenses}
\usepackage[ragged]{sidecap}
\usepackage{bbm}
\usepackage{colortbl}
\usepackage{adjustbox}
\usepackage{mdframed}
\usepackage{listings}
\usepackage[linesnumbered, ruled, vlined]{algorithm2e}
\usepackage{lscape, float} 
\usepackage[most]{tcolorbox}
\usepackage{subcaption}

\tcbset{
    promptstyle/.style={
        enhanced,
        width=\linewidth,
        colback=white,
        colframe=black,
        colbacktitle=gray!20,
        coltitle=black,
        rounded corners,
        boxrule=0.5pt,
        drop shadow=black!50!white,
        attach boxed title to top left={
            xshift=-2mm,
            yshift=-2mm
        },
        boxed title style={
            rounded corners,
            size=small,
            colback=gray!20
        }
    },
    replystyleg/.style={
        enhanced,
        width=\linewidth,
        colback=green!15,
        colframe=black,
        colbacktitle=green!30,
        coltitle=black,
        boxrule=0.5pt,
        drop shadow=black!50!white,
        rounded corners,
        sharp corners=north,
        attach boxed title to top right={
            xshift=-2mm,
            yshift=-2mm
        },
        boxed title style={
            rounded corners,
            size=small,
            colback=green!40
        }
    },
    replystyler/.style={
        enhanced,
        width=\linewidth,
        colback=red!15,
        colframe=black,
        colbacktitle=red!40,
        coltitle=black,
        boxrule=0.5pt,
        drop shadow=black!50!white,
        rounded corners,
        sharp corners=north,
        attach boxed title to top right={
            xshift=-2mm,
            yshift=-2mm
        },
        boxed title style={
            rounded corners,
            size=small,
            colback=red!40
        }
    }
}

\newtcolorbox{promptbox}[1][]{
    promptstyle,
    title=Prompt,
    #1
}

\newcommand*\circled[1]{\tikz[baseline=(char.base)]{
            \node[shape=circle,draw,inner sep=.6pt] (char) {#1};}}

\title{On-Device LLMs for Home Assistant: \\Dual Role in Intent Detection and Response Generation}

\author{Rune Birkmose\textsuperscript{*} \hspace{1em} Nathan Mørkeberg Reece\textsuperscript{*} \hspace{1em} Esben Hofstedt Norvin\textsuperscript{*} \\ \textbf{Johannes Bjerva} \hspace{1em} \textbf{Mike Zhang}\textsuperscript{\textdagger} \\
  Aalborg University, Denmark \\
  \textsuperscript{*}\texttt{\{rbirkm20, nreece20, enorvi20\}@student.aau.dk}\hspace{1em}\textsuperscript{\textdagger}\texttt{jjz@cs.aau.dk} \\
}

\begin{document}
\maketitle
\begingroup\renewcommand\thefootnote{*}
\footnotetext{The authors contributed equally to this work.}
\endgroup
\begin{abstract}
This paper investigates whether Large Language Models (LLMs), fine-tuned on synthetic but domain-representative data, can perform the twofold task of (i) slot and intent detection and (ii) natural language response generation for a smart home assistant, while running solely on resource-limited, CPU-only edge hardware. We fine-tune LLMs to produce both JSON action calls and text responses. Our experiments show that 16-bit and 8-bit quantized variants preserve high accuracy on slot and intent detection and maintain strong semantic coherence in generated text, while the 4-bit model, while retaining generative fluency, suffers a noticeable drop in device-service classification accuracy. Further evaluations on noisy human (non-synthetic) prompts and out-of-domain intents confirm the models' generalization ability, obtaining around 80--86\% accuracy. While the average inference time is 5--6 seconds per query---acceptable for one-shot commands but suboptimal for multi-turn dialogue---our results affirm that an on-device LLM can effectively unify command interpretation and flexible response generation for home automation without relying on specialized hardware.
\end{abstract}

\section{Introduction}
Smart home technologies and IoT devices have proliferated in recent years, with an expected rise from 16.6~billion to 18.8~billion connected devices by the end of 2024 \citep{iotanalytics}. Major providers like Amazon, Google, and Apple typically handle speech recognition and intent detection on cloud servers, which raises user concerns about privacy, data ownership, and reliance on proprietary ecosystems \citep{bbc2025siri}. Conventional solutions for home assistants often rely on specialized, domain-specific classifiers for slot and intent detection (SID), paired with templated system responses. While these approaches can be efficient, they can also be rigid, sometimes requiring precisely phrased user inputs and yielding repetitive or unpersonalized answers.

Recent developments in on-device computing—coupled with improvements in model compression and quantization~\cite{liang2021pruning,gholami2022survey,lang2024comprehensive}—have paved the way for smaller yet still capable language models to run on commodity hardware. These models offer privacy benefits and allow customizable local inference with reduced latency. However, deploying a capable model under strict memory and computational constraints remains challenging. Large-scale Transformer-based language models~\citep{vaswani2017attention}, and especially LLMs~\citep{touvron2023llama, dubey2024llama, bai2023qwen, Qwen, groeneveld-etal-2024-olmo}, have demonstrated remarkable proficiency in tasks ranging from question answering to text generation~\citep{arora2024intent, yin2024large}, yet typically demand substantial hardware resources, restricting them to cloud-based services or large compute clusters.

This paper explores whether a smaller, fine-tuned LLM can provide two capabilities essential to a home assistant---accurate recognition of \emph{what} users want (i.e., slot and intent detection), and \emph{natural} textual responses---while running entirely on an edge device with limited CPU and memory. By unifying these tasks into one end-to-end system, we eliminate the need for separate domain-specific classification modules and templated responses, focusing on efficiency, robust language understanding, and strict correctness in JSON action output.

Additionally, we move away from classic SID datasets and other general spoken language understanding benchmarks. Instead, we investigate whether LLMs can be directly applied to digital assistant software. To this end, we take the open-source Home Assistant software\footnote{\url{https://github.com/home-assistant/core}} as our gold standard for evaluation, targeting real-world device-service pairs and actionable JSON outputs.

\paragraph{Contributions.} Our contributions can be summarized as follows:\footnote{We release all our code and models at~\url{https://github.com/Run396/P9}.}
\circled{1} We show that a 0.5B LLM can be fine-tuned to already jointly handle SID \textit{and} response generation with high accuracy. \circled{2} By quantizing the model (from 16-bit to 8-bit and 4-bit), we quantify trade-offs between memory usage, accuracy, and generative fluency on CPU-only edge hardware. \circled {3} We evaluate the approach on synthetic data, human queries, and out-of-domain tasks, confirming robust generalization.

\section{Related Work}

\paragraph{Slot and Intent Detection.}
Traditional approaches to spoken language understanding (SLU) often treat SID separately using domain-specific classification or sequence tagging approaches \citep{zhang2016joint, wang-etal-2018-bi, weld2022survey, Survey, pham-etal-2023-misca}. More recent transformer-based solutions unify both tasks, leveraging contextual embeddings to improve performance~\citep{castellucci2019multi, van-der-goot-etal-2021-masked, stoica2021intent, arora2024intent} with models like BERT~\citep{devlin-etal-2019-bert}. However, many of these solutions still presume tailored sequence labeling datasets or full-size transformer backends. Our work aligns with the shift to more expressive transformer models for SLU, but we push inference to a local environment while also adding dynamic text generation.

\paragraph{Running LLMs on Edge Devices.}
While training large-scale LLMs remains computationally expensive, numerous works explore strategies for \emph{deploying} them on edge hardware. \citet{arxiv2408} propose FPGA-based accelerators to reduce memory overhead for LLM inference. \citet{EdgeShard} distribute an LLM across multiple low-power devices to increase throughput. An empirical footprint study by \citet{Resource_Footprint} shows that even 7B-parameter models can strain embedded hardware if not sufficiently compressed. Our approach uses a much smaller LLM (0.5B--1.5B parameters) plus weight quantization, showing that near-commodity devices with 8GB RAM can handle both intent classification and text generation if the domain is sufficiently specialized.

\section{Methodology}

Our goal is to integrate two core functionalities of a home assistant into a single model: 

\begin{itemize}
    \item \textbf{Slot and Intent Detection:} The model outputs a valid JSON object that maps to a desired service (intent) and device (slot) pair:
\begin{minted}[frame=single,
               framesep=2mm,
               xleftmargin=0pt,
               tabsize=2]{js}
{"service": "light.turn_on",
  "device": "light.living_room",
  "assistant": "Sure, turning on 
    on the living room light."}
\end{minted}
    \item \textbf{Natural Language Generation:} The model also produces a textual response confirming or elaborating on its action, as can be seen in the example above. The text can then be propagated to, e.g., a text-to-speech model.
\end{itemize}

\noindent
Traditional classifiers only handle device-service classification and do not produce any text. For user-facing text, the baseline approach would rely on templated responses.

\begin{table}[t]
\centering
\small
\begin{tabular}{lccc}
\toprule
\textbf{Partition} & \textbf{Train} & \textbf{Test} & \textbf{Total}\\
\midrule
Classification & 23,372 & 5,843 & 29,215 \\
LLM & 33,361 & 2,435 & 35,796 \\
\bottomrule
\end{tabular}
\caption{\textbf{Aggregated Train/Test Splits.} For the classification baseline, 20\% of the original training set was used as test data (after removing multi-intent samples). The LLM used the full synthetic data; 2,435 remain as test.}
\label{tab:data}
\end{table}

\subsection{Data and Pre-processing}
\label{sec:data}

To the best of our knowledge, there is no existing human-curated dataset specifically for the Home Assistant software. Thus, we rely on synthetic data. We use a publicly available synthetic dataset~\citep{acon96}, which consists of 35,840 synthetic examples designed to mimic Home Assistant commands. Each instance consists of:

\begin{itemize}
    \itemsep0em
    \item A \textbf{User Prompt:} e.g., \emph{``Turn on the kitchen light''}, \emph{``Set the thermostat to 22 degrees''}.
    \item One \textbf{Valid JSON Action}, containing the \texttt{service} and \texttt{device} fields corresponding to Home Assistant calls.
    \item A \textbf{Natural Language Response:} e.g., a paraphrase or affirmation of the action taken.
\end{itemize}

A full example can be found in~\cref{fig:data_example} (\cref{app:sec-data-example}).
We stratify the dataset, maintaining the inherent imbalance (some device types and services appear more frequently). There are 38 service labels and 858 device labels. We split into training and test sets as shown in~\cref{tab:data}. The final training set for the LLM includes $\sim$33k examples, and we set aside 2,435 synthetic samples for evaluation. Note that for the classification-based baselines, we split up the train and test set to separately predict \texttt{service} and \texttt{device} instead of as one prediction, ending up with double the test data (5,843 samples; excluding multi-intent examples). The input consists of only the user message and leave the system message out. A more detailed distribution of the data can be found in~\cref{tab:detailed} (\cref{app:distribution}).

\begin{table}[t]
\centering
\small
\begin{tabular}{lrr}
\toprule
\textbf{Model} & \textbf{Accuracy} & \textbf{BERTScore}\\
\midrule
\textbf{Baselines}           & & \\
SVC Classifier               &  & \\
 \hspace{1em} Service        & 76.6  & ---\\
 \hspace{1em} Device         & 45.4  & ---\\
DistilBERT                   &        &    \\
 \hspace{1em} Service        & 98.8  & ---\\
 \hspace{1em} Device         & 47.9  & ---\\
\midrule
Qwen2.5-0.5B (16-bit)        & 98.8 & 0.84\\
Qwen2.5-0.5B (8-bit)         & 98.4 & 0.79\\
Qwen2.5-0.5B (4-bit)         & 81.7 & 0.88\\
\midrule
Qwen2.5-1.5B (16-bit)        & 96.9 & 0.84\\
Qwen2.5-1.5B (8-bit)         & 96.5 & 0.83\\
Qwen2.5-1.5B (4-bit)         & 90.7 & 0.82\\
\bottomrule
\end{tabular}
\caption{\textbf{Slot/Intent Detection and NLG Results on Synthetic Test Data.} Accuracy is based on exact JSON match. BERTScore measures semantic similarity of the generated text vs.\ gold reference.}
\label{tab:results}
\end{table}

\subsection{Models}

\paragraph{Baseline Classifiers.}
We train a Linear SVC from Scikit-Learn~\citep{pedregosa2011scikit} on TF-IDF features of the user prompt. The classifier outputs a concatenated device-service pair, which is then wrapped in JSON. Additionally, we fine-tune DistilBERT~\cite{sanh2019distilbert} for classification. We use the transformers library~\citep{wolf-etal-2020-transformers} for fine-tuning. We train for 1 epoch using a learning rate of 3$\times 10^{-4}$ with the AdamW optimizer, and a batch size of 64 on a NVIDIA A10 (24GB) GPUs. Both models have no generative capability, so user-facing text is templated.

\paragraph{Small Large Language Models.}
We train using a chat-style format with user--assistant pairs. We primarily use the Qwen2.5-0.5B-Instruct model and the Qwen2.5-1.5B-Instruct~\citep{Qwen}. We fine-tune both models for one epoch with a batch size of 4, using the AdamW optimizer at a learning rate of $2\times10^{-5}$ with a cosine scheduler. The maximum sequence length is set to 2,048 tokens. We use the HuggingFace Transformers library~\citep{wolf-etal-2020-transformers} for training on NVIDIA L4 (24\,GB) GPUs.

\paragraph{Quantization.}
After fine-tuning and having the original 16-bit model, we produce two quantized versions of each model: NF8 and NF4~\cite{dettmers2024qlora}, using bitsandbytes.\footnote{See~\url{https://github.com/bitsandbytes-foundation/bitsandbytes}. We also use double quantization.} 
This allows us to compare accuracy, generative quality, and inference speed under varying memory constraints.

\begin{table}[t]
\centering
\small
\begin{tabular}{lrrr}
\toprule
\textbf{Model}  & \textbf{CPU} & \textbf{T/Q (s)} & \textbf{Load (s)}\\
\midrule
\textbf{Baselines}    & & & \\
SVC Classifier          & 4 & $<$1 & ---\\
DistilBERT              & 4 & $<$1 & ---\\
\midrule
Qwen2.5-0.5B (16-bit)   & 4 & 6.25 & $\pm$3.2 \\
Qwen2.5-1.5B (16-bit)   & 4 & 10.81 & $\pm$5.6 \\
Qwen2.5-0.5B (8-bit)    & 4 & 5.50 & $\pm$3.2 \\
Qwen2.5-1.5B (8-bit)    & 4 & 10.32 & $\pm$5.6 \\
\midrule
Qwen2.5-0.5B (16-bit)   & 2 & 8.49 & $\pm$5.6 \\
Qwen2.5-1.5B (16-bit)   & 2 & 17.72 & $\pm$5.6 \\
Qwen2.5-0.5B (8-bit)    & 2 & 7.89 & $\pm$5.6 \\
Qwen2.5-1.5B (8-bit)    & 2 & 16.11 & $\pm$5.6 \\
\bottomrule
\end{tabular}
\caption{\textbf{Computation Time.} Mean time per query (T/Q) across 500 samples under different CPU core counts and quantization levels. Load time is model initialization.}
\label{tab:performance}
\end{table}

\subsection{Evaluation}
\label{sec:evaluation}
\paragraph{Slot-Intent Detection Accuracy.}
SID must be correct with near-exact string matching, as JSON calls are consumed downstream by the home automation system. We thus parse the model output for the \texttt{service} and \texttt{device} fields; if they match the gold annotation exactly, it is counted as correct. Any mismatch or invalid JSON results in an error.

For the classification task, instead, we separately predict \texttt{service} and \texttt{device} using the same classification model and take the average accuracy.

\paragraph{Text Generation Quality.}
For the natural language responses using the LLMs, we compare each generated response to the reference using BERTScore~\citep{bert-score}.

\paragraph{Inference Environment.}
We simulate a CPU-only setup on an 8\,GB RAM device with up to four CPU cores. We measure average inference time on a 500-sample subset, varying both quantization level and the number of CPU cores.

\begin{table}[t]
\centering
\small
\begin{tabular}{lrr}
\toprule
\textbf{Model} & \textbf{Accuracy} & \textbf{BERTScore}\\
\midrule
Qwen2.5-0.5B    &   80.0    &   0.76 \\
Qwen2.5-1.5B    &   86.7    &   0.74 \\
\bottomrule
\end{tabular}
\caption{\textbf{Results Out-of-Domain Queries.} Accuracy and BERTScore over 60 OOD samples.}
\label{tab:ood_eval}
\end{table}

\section{Results}

\subsection{Slot and Intent Detection}
\cref{tab:results} shows the SID performance of both the 0.5B and 1.5B LLMs under various quantization levels, alongside the baseline SVC and DistilBERT. For the 0.5B model, the 16-bit and 8-bit variants reach near-perfect accuracy ($\sim 99\%$). The 4-bit version drops to 81.7\%, which is still better than the the SVC baseline (average 61.0\% accuracy) and DistilBERT baseline (average 73.4\% accuracy).

Interestingly, for the larger 1.5B model, the 16-bit and 8-bit variants achieve 96.9\% and 96.5\% accuracy, respectively, while the 4-bit version gets 90.7\%. Thus, while the smaller 0.5B model actually yields higher raw accuracy in-domain, the 1.5B model remains competitive and in some out-of-domain tests (next section) performs better.

\subsection{Natural Language Generation}
Although the 4-bit models suffer in SID accuracy, \cref{tab:results} shows that the 0.5B 4-bit variant has the highest BERTScore (0.88). This indicates that while it may misclassify device/service fields, the generative text can still be fluent and semantically close to the target. Meanwhile, the 8-bit versions drop in BERTScore for the 0.5B model (0.79) and remain steady for the 1.5B model (0.83). Qualitative samples show that small changes in quantization can shift the style and lexical choices of the generated text.

\subsection{Inference Time and Memory}

\cref{tab:performance} summarizes the inference speed across model size, quantization, and CPU core settings. The 8-bit model is only slightly faster than the 16-bit model (5.5\,s vs.\ 6.25\,s on 4 cores for the 0.5B). Doubling CPU cores from 2 to 4 reduces latency roughly by half. The 1.5B model takes longer (up to 10--17\,s per query), which may be borderline for real-time usage in multi-turn dialogues.

\begin{table}[t]
\centering
\small
\begin{tabular}{lrr}
\toprule
\textbf{Model} & \textbf{Accuracy} & \textbf{BERTScore}\\
\midrule
Qwen2.5-0.5B    &   84.0    &   0.68 \\
Qwen2.5-1.5B    &   86.4    &   0.66 \\
\bottomrule
\end{tabular}
\caption{\textbf{Results Human-Generated Queries.} Accuracy and BERTScore over 81 real-user queries.}
\label{tab:human_eval}
\end{table}

\subsection{Out-of-Domain Intents}

In \cref{tab:ood_eval}, we evaluate 60 OOD queries that mention either novel device types or services not appearing in the training set. The 0.5B model scores 80.0\% accuracy vs.\ 86.7\% for the 1.5B model, with BERTScores of 0.76 and 0.74 respectively. The results suggest that the 1.5B model generalizes somewhat better to unfamiliar domains, though both degrade compared to in-domain performance.

\subsection{Human Prompts}
Finally, we tested each model on 81 human-written prompts. Ten participants (ages 23--69) contributed typical commands they would issue to a home assistant, including incomplete or ambiguous phrasing. \cref{tab:human_eval} shows that the 0.5B model achieves 84.0\% accuracy, whereas the 1.5B model is slightly higher at 86.4\%. BERTScores are around 0.66--0.68. The gap vs.\ synthetic data reflects real-user queries with more variation and noisy data.

\section{Discussion}
Despite near-perfect performance on the synthetic test set,~\cref{tab:ood_eval} and~\ref{tab:human_eval} reveal a drop to 80--86\% accuracy in real or out-of-domain queries. This discrepancy likely stems from the difficulty of handling spontaneous human phrasing, missing location or device details, and genuinely novel device types. Still, the results surpass the SVC and DistilBERT baseline.

Interestingly, while the 4-bit model can generate fluent natural language responses (often scoring the highest BERTScore in the 0.5B case), its classification accuracy suffers. This underscores that quantizing a model to extreme levels can degrade structured predictions more than open-ended text generation.

Regarding speed, the 1.5B model yields consistent accuracy gains on OOD data but also increases inference time by up to 2--3$\times$. For single-turn commands, 5--6 seconds per query might be acceptable, but multi-turn dialogue would require faster or more efficient strategies. Future work may explore parameter-efficient fine-tuning, context truncation, or advanced quantization (e.g., 8-bit + partial 4-bit layering) to reduce inference times.

\section{Conclusion}
We present that LLMs can simultaneously perform SID and natural language response generation for a home automation domain. Experiments on an 8GB RAM, CPU-only environment show that 8-bit quantization largely preserves in-domain accuracy (up to 99\%) and strong text fluency, while 4-bit introduces significant classification errors despite retaining good generative capability. We further demonstrate promising generalization to human-written prompts and out-of-domain tasks, with accuracy around 80--86\%. However, per-query inference times of 5--6 seconds indicate that LLM-based assistants, as implemented here, are not yet ideal for fast multi-turn dialogues on edge devices. Future work can refine these models for faster, more memory-efficient inference, enabling privacy-preserving yet flexible home automation assistants.

\section*{Limitations} 
Our use of synthetic data may limit the diversity of user prompts; while we partially mitigated this with human-written queries, data coverage remains a challenge. The model also relies on structurally valid JSON output. Real-world usage may need fallback logic to handle malformed or incomplete responses. Moreover, we focus on a single domain (home automation); scaling to broader or open-ended tasks likely requires larger models and may degrade performance under CPU-only constraints.

\section*{Ethical Considerations}
We do not foresee any major ethical issues with this work. The primary domain is home automation, and the dataset is synthetic or user-provided under informed consent. Nonetheless, deploying generative models in user-facing applications requires caution regarding hallucinated or incorrect responses, as well as user data privacy.

\section*{Acknowledgments}
MZ and JB are supported by the research grant (VIL57392) from VILLUM FONDEN.

\bibliography{custom, anthology}

\begin{thebibliography}{31}
\providecommand{\natexlab}[1]{#1}

\bibitem[{{acon96}(2024)}]{acon96}
{acon96}. 2024.
\newblock {Home Assistant Requests Dataset}.
\newblock \url{https://huggingface.co/datasets/acon96/Home-Assistant-Requests}.
\newblock Accessed: 2025.

\bibitem[{Arora et~al.(2024)Arora, Jain, and Merugu}]{arora2024intent}
Gaurav Arora, Shreya Jain, and Srujana Merugu. 2024.
\newblock \href {https://arxiv.org/pdf/2410.01627} {Intent detection in the age of llms}.
\newblock \emph{arXiv preprint arXiv:2410.01627}.

\bibitem[{Bai et~al.(2023)Bai, Bai, Chu, Cui, Dang, Deng, Fan, Ge, Han, Huang et~al.}]{bai2023qwen}
Jinze Bai, Shuai Bai, Yunfei Chu, Zeyu Cui, Kai Dang, Xiaodong Deng, Yang Fan, Wenbin Ge, Yu~Han, Fei Huang, et~al. 2023.
\newblock \href {https://arxiv.org/abs/2309.16609} {Qwen technical report}.
\newblock \emph{ArXiv preprint}, abs/2309.16609.

\bibitem[{{BBC News}(2025)}]{bbc2025siri}
{BBC News}. 2025.
\newblock \href {https://www.bbc.com/news/articles/cr4rvr495rgo} {Apple to pay \$95m to settle siri 'listening' lawsuit}.

\bibitem[{Castellucci et~al.(2019)Castellucci, Bellomaria, Favalli, and Romagnoli}]{castellucci2019multi}
Giuseppe Castellucci, Valentina Bellomaria, Andrea Favalli, and Raniero Romagnoli. 2019.
\newblock Multi-lingual intent detection and slot filling in a joint bert-based model.
\newblock \emph{arXiv preprint arXiv:1907.02884}.

\bibitem[{Dettmers et~al.(2024)Dettmers, Pagnoni, Holtzman, and Zettlemoyer}]{dettmers2024qlora}
Tim Dettmers, Artidoro Pagnoni, Ari Holtzman, and Luke Zettlemoyer. 2024.
\newblock Qlora: Efficient finetuning of quantized llms.
\newblock \emph{Advances in Neural Information Processing Systems}, 36.

\bibitem[{Devlin et~al.(2019)Devlin, Chang, Lee, and Toutanova}]{devlin-etal-2019-bert}
Jacob Devlin, Ming-Wei Chang, Kenton Lee, and Kristina Toutanova. 2019.
\newblock \href {https://doi.org/10.18653/v1/N19-1423} {{BERT}: Pre-training of deep bidirectional transformers for language understanding}.
\newblock In \emph{Proceedings of the 2019 Conference of the North {A}merican Chapter of the Association for Computational Linguistics: Human Language Technologies, Volume 1 (Long and Short Papers)}, pages 4171--4186, Minneapolis, Minnesota. Association for Computational Linguistics.

\bibitem[{Dhar et~al.(2024)Dhar, Deng, Lo, Wu, Zhao, and Suo}]{Resource_Footprint}
Nobel Dhar, Bobin Deng, Dan Lo, Xiaofeng Wu, Liang Zhao, and Kun Suo. 2024.
\newblock \href {https://doi.org/10.1145/3603287.3651205} {An empirical analysis and resource footprint study of deploying large language models on edge devices}.
\newblock \emph{Proceedings of the ACM}.

\bibitem[{Dubey et~al.(2024)Dubey, Jauhri, Pandey, Kadian, Al-Dahle, Letman, Mathur, Schelten, Yang, Fan et~al.}]{dubey2024llama}
Abhimanyu Dubey, Abhinav Jauhri, Abhinav Pandey, Abhishek Kadian, Ahmad Al-Dahle, Aiesha Letman, Akhil Mathur, Alan Schelten, Amy Yang, Angela Fan, et~al. 2024.
\newblock \href {https://arxiv.org/abs/2407.21783} {The llama 3 herd of models}.
\newblock \emph{ArXiv preprint}, abs/2407.21783.

\bibitem[{Gholami et~al.(2022)Gholami, Kim, Dong, Yao, Mahoney, and Keutzer}]{gholami2022survey}
Amir Gholami, Sehoon Kim, Zhen Dong, Zhewei Yao, Michael~W Mahoney, and Kurt Keutzer. 2022.
\newblock A survey of quantization methods for efficient neural network inference.
\newblock In \emph{Low-Power Computer Vision}, pages 291--326. Chapman and Hall/CRC.

\bibitem[{Groeneveld et~al.(2024)Groeneveld, Beltagy, Walsh, Bhagia, Kinney, Tafjord, Jha, Ivison, Magnusson, Wang, Arora, Atkinson, Authur, Chandu, Cohan, Dumas, Elazar, Gu, Hessel, Khot, Merrill, Morrison, Muennighoff, Naik, Nam, Peters, Pyatkin, Ravichander, Schwenk, Shah, Smith, Strubell, Subramani, Wortsman, Dasigi, Lambert, Richardson, Zettlemoyer, Dodge, Lo, Soldaini, Smith, and Hajishirzi}]{groeneveld-etal-2024-olmo}
Dirk Groeneveld, Iz~Beltagy, Evan Walsh, Akshita Bhagia, Rodney Kinney, Oyvind Tafjord, Ananya Jha, Hamish Ivison, Ian Magnusson, Yizhong Wang, Shane Arora, David Atkinson, Russell Authur, Khyathi Chandu, Arman Cohan, Jennifer Dumas, Yanai Elazar, Yuling Gu, Jack Hessel, Tushar Khot, William Merrill, Jacob Morrison, Niklas Muennighoff, Aakanksha Naik, Crystal Nam, Matthew Peters, Valentina Pyatkin, Abhilasha Ravichander, Dustin Schwenk, Saurabh Shah, William Smith, Emma Strubell, Nishant Subramani, Mitchell Wortsman, Pradeep Dasigi, Nathan Lambert, Kyle Richardson, Luke Zettlemoyer, Jesse Dodge, Kyle Lo, Luca Soldaini, Noah Smith, and Hannaneh Hajishirzi. 2024.
\newblock \href {https://doi.org/10.18653/v1/2024.acl-long.841} {{OLM}o: Accelerating the science of language models}.
\newblock In \emph{Proceedings of the 62nd Annual Meeting of the Association for Computational Linguistics (Volume 1: Long Papers)}, pages 15789--15809, Bangkok, Thailand. Association for Computational Linguistics.

\bibitem[{Haris et~al.(2024)Haris, Saha, Hu, and Cano}]{arxiv2408}
Jude Haris, Rappy Saha, Wenhao Hu, and José Cano. 2024.
\newblock \href {https://arxiv.org/pdf/2408.00462} {Designing efficient llm accelerators for edge devices}.
\newblock \emph{arXiv preprint arXiv:2408.00462}.

\bibitem[{{IoT Analytics}(2024)}]{iotanalytics}
{IoT Analytics}. 2024.
\newblock {Number of Connected IoT Devices}.
\newblock \url{https://iot-analytics.com/number-connected-iot-devices/}.
\newblock Accessed: 2024-10-16.

\bibitem[{Lang et~al.(2024)Lang, Guo, and Huang}]{lang2024comprehensive}
Jiedong Lang, Zhehao Guo, and Shuyu Huang. 2024.
\newblock A comprehensive study on quantization techniques for large language models.
\newblock \emph{arXiv preprint arXiv:2411.02530}.

\bibitem[{Liang et~al.(2021)Liang, Glossner, Wang, Shi, and Zhang}]{liang2021pruning}
Tailin Liang, John Glossner, Lei Wang, Shaobo Shi, and Xiaotong Zhang. 2021.
\newblock Pruning and quantization for deep neural network acceleration: A survey.
\newblock \emph{Neurocomputing}, 461:370--403.

\bibitem[{Pedregosa et~al.(2011)Pedregosa, Varoquaux, Gramfort, Michel, Thirion, Grisel, Blondel, Prettenhofer, Weiss, Dubourg et~al.}]{pedregosa2011scikit}
Fabian Pedregosa, Ga{\"e}l Varoquaux, Alexandre Gramfort, Vincent Michel, Bertrand Thirion, Olivier Grisel, Mathieu Blondel, Peter Prettenhofer, Ron Weiss, Vincent Dubourg, et~al. 2011.
\newblock Scikit-learn: Machine learning in python.
\newblock \emph{the Journal of machine Learning research}, 12:2825--2830.

\bibitem[{Pham et~al.(2023)Pham, Tran, and Nguyen}]{pham-etal-2023-misca}
Thinh Pham, Chi Tran, and Dat~Quoc Nguyen. 2023.
\newblock \href {https://doi.org/10.18653/v1/2023.findings-emnlp.841} {{MISCA}: A joint model for multiple intent detection and slot filling with intent-slot co-attention}.
\newblock In \emph{Findings of the Association for Computational Linguistics: EMNLP 2023}, pages 12641--12650, Singapore. Association for Computational Linguistics.

\bibitem[{Qin et~al.(2021)Qin, Xie, Che, and Liu}]{Survey}
Libo Qin, Tianbao Xie, Wanxiang Che, and Ting Liu. 2021.
\newblock \href {https://arxiv.org/abs/2103.03095} {A survey on spoken language understanding: Recent advances and new frontiers}.
\newblock \emph{arXiv preprint arXiv:2103.03095}.

\bibitem[{Sanh et~al.(2019)Sanh, Debut, Chaumond, and Wolf}]{sanh2019distilbert}
Victor Sanh, Lysandre Debut, Julien Chaumond, and Thomas Wolf. 2019.
\newblock Distilbert, a distilled version of bert: smaller, faster, cheaper and lighter.
\newblock \emph{arXiv preprint arXiv:1910.01108}.

\bibitem[{Stoica et~al.(2021)Stoica, Kadar, Lemnaru, Potolea, and D{\^\i}n{\c{s}}oreanu}]{stoica2021intent}
Anda Stoica, Tibor Kadar, Camelia Lemnaru, Rodica Potolea, and Mihaela D{\^\i}n{\c{s}}oreanu. 2021.
\newblock Intent detection and slot filling with capsule net architectures for a romanian home assistant.
\newblock \emph{Sensors}, 21(4):1230.

\bibitem[{Touvron et~al.(2023)Touvron, Martin, Stone, Albert, Almahairi, Babaei, Bashlykov, Batra, Bhargava, Bhosale et~al.}]{touvron2023llama}
Hugo Touvron, Louis Martin, Kevin Stone, Peter Albert, Amjad Almahairi, Yasmine Babaei, Nikolay Bashlykov, Soumya Batra, Prajjwal Bhargava, Shruti Bhosale, et~al. 2023.
\newblock \href {https://arxiv.org/abs/2307.09288} {Llama 2: Open foundation and fine-tuned chat models}.
\newblock \emph{ArXiv preprint}, abs/2307.09288.

\bibitem[{van~der Goot et~al.(2021)van~der Goot, Sharaf, Imankulova, {\"U}st{\"u}n, Stepanovi{\'c}, Ramponi, Khairunnisa, Komachi, and Plank}]{van-der-goot-etal-2021-masked}
Rob van~der Goot, Ibrahim Sharaf, Aizhan Imankulova, Ahmet {\"U}st{\"u}n, Marija Stepanovi{\'c}, Alan Ramponi, Siti~Oryza Khairunnisa, Mamoru Komachi, and Barbara Plank. 2021.
\newblock \href {https://doi.org/10.18653/v1/2021.naacl-main.197} {From masked language modeling to translation: Non-{E}nglish auxiliary tasks improve zero-shot spoken language understanding}.
\newblock In \emph{Proceedings of the 2021 Conference of the North American Chapter of the Association for Computational Linguistics: Human Language Technologies}, pages 2479--2497, Online. Association for Computational Linguistics.

\bibitem[{Vaswani et~al.(2017)Vaswani, Shazeer, Parmar, Uszkoreit, Jones, Gomez, Kaiser, and Polosukhin}]{vaswani2017attention}
A~Vaswani, N~Shazeer, N~Parmar, J~Uszkoreit, L~Jones, A~Gomez, L~Kaiser, and I~Polosukhin. 2017.
\newblock {Attention is All You Need}.
\newblock In \emph{Advances in Neural Information Processing Systems (NIPS)}.

\bibitem[{Wang et~al.(2018)Wang, Shen, and Jin}]{wang-etal-2018-bi}
Yu~Wang, Yilin Shen, and Hongxia Jin. 2018.
\newblock \href {https://doi.org/10.18653/v1/N18-2050} {A bi-model based {RNN} semantic frame parsing model for intent detection and slot filling}.
\newblock In \emph{Proceedings of the 2018 Conference of the North {A}merican Chapter of the Association for Computational Linguistics: Human Language Technologies, Volume 2 (Short Papers)}, pages 309--314, New Orleans, Louisiana. Association for Computational Linguistics.

\bibitem[{Weld et~al.(2022)Weld, Huang, Long, Poon, and Han}]{weld2022survey}
Henry Weld, Xiaoqi Huang, Siqu Long, Josiah Poon, and Soyeon~Caren Han. 2022.
\newblock A survey of joint intent detection and slot filling models in natural language understanding.
\newblock \emph{ACM Computing Surveys}, 55(8):1--38.

\bibitem[{Wolf et~al.(2020)Wolf, Debut, Sanh, Chaumond, Delangue, Moi, Cistac, Rault, Louf, Funtowicz, Davison, Shleifer, von Platen, Ma, Jernite, Plu, Xu, Le~Scao, Gugger, Drame, Lhoest, and Rush}]{wolf-etal-2020-transformers}
Thomas Wolf, Lysandre Debut, Victor Sanh, Julien Chaumond, Clement Delangue, Anthony Moi, Pierric Cistac, Tim Rault, Remi Louf, Morgan Funtowicz, Joe Davison, Sam Shleifer, Patrick von Platen, Clara Ma, Yacine Jernite, Julien Plu, Canwen Xu, Teven Le~Scao, Sylvain Gugger, Mariama Drame, Quentin Lhoest, and Alexander Rush. 2020.
\newblock \href {https://doi.org/10.18653/v1/2020.emnlp-demos.6} {Transformers: State-of-the-art natural language processing}.
\newblock In \emph{Proceedings of the 2020 Conference on Empirical Methods in Natural Language Processing: System Demonstrations}, pages 38--45, Online. Association for Computational Linguistics.

\bibitem[{Yang et~al.(2024)Yang, Yang, and Zhang}]{Qwen}
An~Yang, Baosong Yang, and Beichen Zhang. 2024.
\newblock \href {https://arxiv.org/abs/2412.15115} {{Qwen2.5 Technical Report}}.
\newblock \emph{arXiv preprint arXiv:2412.15115}.

\bibitem[{Yin et~al.(2024)Yin, Huang, Xu, Huang, and Chen}]{yin2024large}
Shangjian Yin, Peijie Huang, Yuhong Xu, Haojing Huang, and Jiatian Chen. 2024.
\newblock Do large language models understand multi-intent spoken language?
\newblock \emph{arXiv preprint arXiv:2403.04481}.

\bibitem[{Zhang et~al.(2024)Zhang, Cao, Shen, and Cui}]{EdgeShard}
Mingjin Zhang, Jiannong Cao, Xiaoming Shen, and Zeyang Cui. 2024.
\newblock \href {https://arxiv.org/abs/2405.14371} {Edgeshard: Efficient llm inference via collaborative edge computing}.
\newblock \emph{arXiv preprint arXiv:2405.14371}.

\bibitem[{Zhang et~al.(2020)Zhang, Kishore, Wu, Weinberger, and Artzi}]{bert-score}
Tianyi Zhang, Varsha Kishore, Felix Wu, Kilian~Q. Weinberger, and Yoav Artzi. 2020.
\newblock \href {https://openreview.net/forum?id=SkeHuCVFDr} {{BERTScore}: Evaluating text generation with {BERT}}.
\newblock In \emph{International Conference on Learning Representations}.

\bibitem[{Zhang and Wang(2016)}]{zhang2016joint}
Xiaodong Zhang and Houfeng Wang. 2016.
\newblock A joint model of intent determination and slot filling for spoken language understanding.
\newblock In \emph{Proceedings of the International Joint Conference on Artificial Intelligence (IJCAI)}, volume~16, pages 2993--2999.

\end{thebibliography}

\clearpage

\appendix

\section{Data Examples}
\label{app:sec-data-example}
We show a data example in~\cref{fig:data_example}.

\begin{figure*}
\begin{minted}[frame=single,framesep=2mm,xleftmargin=0pt,tabsize=2]{js}
[ 
    { 
      "from": "system", 
      "value": "You are 'Al', a helpful AI Assistant that controls the devices in 
      a house. Complete the following task as instructed or answer the following 
      question with the information provided only.\nServices: cover.close_cover(), 
      cover.open_cover(), cover.stop_cover(), cover.toggle(), lock.lock(), 
      lock.unlock(), media_player.media_next_track(), media_player.media_pause(), 
      media_player.media_play(), media_player.media_play_pause(), 
      media_player.media_previous_track(), media_player.media_stop(), 
      media_player.toggle(), media_player.turn_off(), media_player.turn_on(), 
      media_player.volume_down(), media_player.volume_mute(), 
      media_player.volume_up(), switch.toggle(), switch.turn_off(), 
      switch.turn_on(), timer.cancel(), timer.pause(), timer.start(duration), 
      vacuum.pause(), vacuum.return_to_base(), vacuum.start(), vacuum.stop()
      Devices: media_player.harman_kardon_aura 'Harman Kardon Glass Speaker' 
      = standby; vol=0.88\ntimer.kitchen_oven 'Kitchen oven timer' = 
      active\nlock.office_cabinet 'Office cabinet lock' = unlocked
      cover.master_bedroom 'Master Bedroom' = closed
      vacuum.hallway_neato 'Hallway path cleaner' = docked
      switch.basement_lights 'Basement Lights Switch' = off" 
    }, 
    { 
      "from": "user", 
      "value": "reverse the master bedroom blinds" 
    }, 
    { 
      "from": "assistant", 
      "value": "switching Master Bedroom state as requested\n
      ```homeassistant
         {
          \"service\": \"cover.toggle\", 
          \"target_device\": \"cover.master_bedroom\"
        }
      ```" 
    }
]
\end{minted}
\caption{\textbf{Data Example.} In the figure, we show a data example from the~\citet{acon96} dataset.}
\label{fig:data_example}
\end{figure*}

\clearpage
\section{Data Distribution Detailed}
\label{app:distribution}

We show a more detailed distribution of the dataset in~\cref{tab:detailed}.

\begin{table*}[t]
\centering
\begin{tabular}{lrr}
\toprule
\textbf{Class} & \textbf{Total Dataset} & \textbf{Test} \\
\midrule
climate.set\_fan\_mode        & 1080  & 0    \\
climate.set\_humidity         & 1080  & 0    \\
climate.set\_hvac\_mode       & 1080  & 0    \\
climate.set\_temperature      & 1000  & 0    \\
cover.close                   & 385   & 35   \\
cover.open                    & 395   & 40   \\
cover.stop                    & 320   & 25   \\
cover.toggle                  & 365   & 25   \\
fan.decrease\_speed           & 360   & 60   \\
fan.increase\_speed           & 300   & 40   \\
fan.toggle                    & 390   & 85   \\
fan.turn\_off                 & 390   & 70   \\
fan.turn\_on                  & 405   & 60   \\
light.toggle                  & 450   & 90   \\
light.turn\_off               & 2535  & 600  \\
light.turn\_on                & 11940 & 150  \\
lock.lock                     & 200   & 125  \\
lock.unlock                   & 185   & 125  \\
media\_player.media\_next\_track & 55    & 25   \\
media\_player.media\_pause       & 55    & 25   \\
media\_player.media\_play        & 70    & 25   \\
media\_player.media\_previous\_track & 55 & 25 \\
media\_player.media\_stop        & 55    & 25   \\
media\_player.turn\_off          & 25    & 25   \\
media\_player.turn\_on           & 40    & 40   \\
media\_player.volume\_down       & 65    & 35   \\
media\_player.volume\_mute       & 60    & 30   \\
media\_player.volume\_up         & 85    & 40   \\
switch.toggle                 & 250   & 50   \\
switch.turn\_off              & 500   & 175  \\
switch.turn\_on               & 540   & 165  \\
timer.cancel                  & 600   & 0    \\
timer.start                   & 600   & 0    \\
todo.add\_item                & 1560  & 0    \\
vacuum.pause                  & 15    & 0    \\
vacuum.return\_to\_base       & 150   & 0    \\
vacuum.start                  & 370   & 220  \\
vacuum.stop                   & 15    & 0    \\
\bottomrule
\end{tabular}
\caption{\textbf{Detailed Class Distribution Service.} Total Dataset vs.\ LLM Test Subset}
\label{tab:detailed}
\end{table*}

\end{document}